# Learning Link-Probabilities in Causal Trees

by
Igor Roizen & Judea Pearl
UCLA Computer Science Department
Cognitive Systems Laboratory

**Abstract**

A learning algorithm is presented which given the structure of a causal tree, will estimate its link probabilities by sequential measurements on the leaves only. Internal nodes of the tree represent conceptual (hidden) variables inaccessible to observation. The method described is incremental, local, efficient, and remains robust to measurement imprecisions.

## 1  The Problem

A *causal tree* offers both a natural data structure for expressing empirical knowledge and an effective computational scheme for reasoning about the interpretation of sensory information [Pearl 1985]. In causal trees, each node represents a binary random variable, and every directed link $w \to z$ signifies a direct dependency between $w$ and $z$, its strength quantified by a conditional probability matrix with entries $P(z|w)$, $w = 0, 1$, $z = 0, 1$. Each matrix contains two independent parameters,

$$f \triangleq P(z=1|w=1) \quad \text{and} \quad g \triangleq P(z=1|w=0).$$

These parameters, when given for all links along with the prior probability $\alpha$ of the root $r$, completely define a joint probability distribution of all the node variables involved, which reflects the topology of the tree.

This paper addresses the following learning task. We imagine that the leaves of a causal tree represent observable (visible) variables while the intermediate nodes represent conceptual (hidden) variables, inaccessible to direct observation. An agent is given only the structure of the tree, without

211

the link probabilities, and is permitted to conduct sequential measurements on the leaves only. Each measurement consists of a sample of leaf values, $D = (x_1, x_2, \ldots, x_n)$, drawn independently from the distribution defined by the tree. The task is to estimate the link-probabilities for *all* the links, peripheral as well as internal.

If we could measure the sampled values of $w$ and $z$ for each link $w \to z$, then both $f$ and $g$ can be estimated by traditional techniques such as counting, incrementing, and Bayes updating [Spiegelhalter 1986]. However, the fact that these measurements do not expose the actual values of the intermediate variables renders this learning task "unsupervised," requiring novel treatment. This paper introduces an effective learning algorithm for updating the estimates of all link probabilities. Its effectiveness is due primarily to the efficiency and autonomy with which beliefs can be propagated along causal trees [Pearl 1982].

## 2 The Learning Algorithm

The learning algorithm developed is based on the observation that, for each link $w \to z$,
$$P(w|z) = \sum_D P(w|z, D) P(D).$$

Thus, if we obtain an estimate of $P(w|z, D)$ for a given sample vector $D = (x_1, x_2, \ldots, x_n)$, we can also obtain an estimate of $P(w|z)$ by averaging the former over all of the observed samples. Moreover, with an estimate of $P(w|z)$ at hand, the desired link probabilities $P(z|w)$ can be calculated from Bayes rule $P(z|w) = P(w|z)P(z)/P(w)$, using the computed estimates of $P(z)$ and $P(w)$. To facilitate this, we define the *belief* function

$$Bel(w) \triangleq P(w|D)$$

and compute, for each observation $D$, the estimates $\hat{Bel}(w|z)$, $\hat{Bel}(z)$ and $\hat{Bel}(w)$, using the propagation scheme described in [Pearl 1982] and the current estimated parameters, $\hat{\alpha}$, $\hat{f}$'s and $\hat{g}$'s. Averaging $\hat{Bel}(w|z)$, $\hat{Bel}(z)$, and $\hat{Bel}(w)$ over a sequence of samples provides estimates for $P(w|z), P(z)$, and $P(w)$, respectively, and permits updating of the parameters $\hat{\alpha}$, $\hat{f}$ and $\hat{g}$, according to their probabilistic definitions.

Let $Avg[u(D)]$ be the sample average of a function $u(\cdot)$ of a random sample $D$, computed after the arrival of the $n$-th sample, namely:
$$Avg\left[u(D_n)\right] = \frac{(n-1)\,Avg\left[u(D_{n-1})\right]\ +\ u(D_n)}{n}.$$
Our learning algorithm can then be described as follows.

- Initialize the link parameters to the best currently available estimates.
- Repeat the steps below until the desired tolerance is attained:

  1. Observe a new sample vector $D$.

  2. For each link $w \to z$, where $z$ is non-terminal, update the estimates
  $$\hat{P}(w=1) := Avg\left[\hat{Bel}(w=1)\right],$$
  $$\hat{P}(w=1|z=1) := Avg\left[\hat{Bel}(w=1|z=1)\right],$$
  $$\hat{P}(w=1|z=0) := Avg\left[\hat{Bel}(w=1|z=0)\right].$$

  For links $w \to z$ with $z$ terminal, update $\hat{P}(w=1)$ and $\hat{P}(z=1) := Avg(z)$; if the $z$-component of $D$ is 1, update $\hat{P}(w=1|z=1)$, otherwise, update $\hat{P}(w=1|z=0)$.

  3. For each link $w \to z$, update the estimates of the link parameters and the prior probability $\alpha$ as follows:
  $$\hat{f} := \frac{\hat{P}(w=1|z=1)\hat{P}(z=1)}{\hat{P}(w=1)} \cdot \hat{P}(z=1)$$
  $$+ \left\{1 - \frac{\hat{P}(w=1|z=0)\left[1-\hat{P}(z=1)\right]}{\hat{P}(w=1)}\right\} \cdot \left[1 - \hat{P}(z=1)\right],$$
  $$\hat{g} := \frac{\left[1-\hat{P}(w=1|z=1)\right]\hat{P}(z=1)}{1-\hat{P}(w=1)} \cdot \hat{P}(z=1)$$
  $$+ \left\{1 - \frac{\left[1-\hat{P}(w=1|z=0)\right]\left[1-\hat{P}(z=1)\right]}{1-\hat{P}(w=1)}\right\} \cdot \left[1 - \hat{P}(z=1)\right],$$
  $$\hat{\alpha} := \hat{P}(r=1). \qquad \square$$

Each updating equation for $\hat{f}$ and $\hat{g}$ consists of a weighted average of two terms. The first term follows directly from the corresponding definition, the second was introduced in order to utilize $\hat{Bel}(w|z)$ for both $z=1$ and $z=0$, thus making more effective use of the available data.

## 3  Conclusion

Pearl [Pearl 1985] showed that, if the joint distribution of the visible variables is known with absolute precision, both the structure of the tree and the link probabilities can be determined uniquely. His method, however, is non-incremental and extremely sensitive to imprecisions due to the finite number of measurements involved. By contrast, the method described in this paper is incremental, local, and remains robust to measurement imprecisions.

Simulation results show that the estimates converge to their exact values at the same rate as the frequency estimates would, were the values of the internal variables also observed.